\documentclass[sn-mathphys]{sn-jnl}

\usepackage[T1]{fontenc}
\usepackage[latin9]{inputenc}
\usepackage{amsmath}
\usepackage{amsthm}
\usepackage{graphicx}
\usepackage{amsmath,amssymb,amsfonts}
\usepackage{textcomp}
\usepackage{xcolor}
\usepackage{float}
\usepackage{array}
\usepackage{algorithm}
\usepackage[]{algpseudocode}
\usepackage{bbm}
\usepackage{mathtools}
\usepackage{afterpage}
\usepackage{lscape}

\usepackage{framed,multirow}

\usepackage{amssymb}
\usepackage{latexsym}

\usepackage{url}
\usepackage{xcolor}
\usepackage{babel}

\newcolumntype{H}{>{\setbox0=\hbox\bgroup}c<{\egroup}@{}}

\newcommand\ddownarrow{\mathchoice%
    {\rotatebox[origin=c]{-135}{$\displaystyle\dashrightarrow$}}%
    {\rotatebox[origin=c]{-135}{$\displaystyle\dashrightarrow$}}%
    {\rotatebox[origin=c]{-135}{$\scriptstyle\dashrightarrow$}}%
    {\rotatebox[origin=c]{-135}{$\scriptscriptstyle\dashrightarrow$}}%
}

\newcommand\duparrow{\mathchoice%
    {\rotatebox[origin=c]{135}{$\displaystyle\dashrightarrow$}}%
    {\rotatebox[origin=c]{135}{$\displaystyle\dashrightarrow$}}%
    {\rotatebox[origin=c]{135}{$\scriptstyle\dashrightarrow$}}%
    {\rotatebox[origin=c]{135}{$\scriptscriptstyle\dashrightarrow$}}%
}

\newcommand\ndownarrow{\mathchoice%
  {\rotatebox[origin=c]{-45}{$\displaystyle\downarrow$}}
  {\rotatebox[origin=c]{-45}{$\displaystyle\downarrow$}}
  {\rotatebox[origin=c]{-45}{$\scriptstyle\downarrow$}}
  {\rotatebox[origin=c]{-45}{$\scriptscriptstyle\downarrow$}}
}

\newcommand\nuparrow{\mathchoice%
  {\rotatebox[origin=c]{45}{$\displaystyle\uparrow$}}
  {\rotatebox[origin=c]{45}{$\displaystyle\uparrow$}}
  {\rotatebox[origin=c]{45}{$\scriptstyle\uparrow$}}
  {\rotatebox[origin=c]{45}{$\scriptscriptstyle\uparrow$}}
}

\newcommand\nDownarrow{\mathchoice%
  {\rotatebox[origin=c]{-45}{$\displaystyle\Downarrow$}}
  {\rotatebox[origin=c]{-45}{$\displaystyle\Downarrow$}}
  {\rotatebox[origin=c]{-45}{$\scriptstyle\Downarrow$}}
  {\rotatebox[origin=c]{-45}{$\scriptscriptstyle\Downarrow$}}
}

\newcommand\nUparrow{\mathchoice%
  {\rotatebox[origin=c]{45}{$\displaystyle\Uparrow$}}
  {\rotatebox[origin=c]{45}{$\displaystyle\Uparrow$}}
  {\rotatebox[origin=c]{45}{$\scriptstyle\Uparrow$}}
  {\rotatebox[origin=c]{45}{$\scriptscriptstyle\Uparrow$}}
}

\DeclarePairedDelimiter\ceil{\lceil}{\rceil}

\providecommand{\tabularnewline}{\\}

\jyear{2021}%

\theoremstyle{thmstyleone}%
\theoremstyle{thmstyletwo}%

\theoremstyle{thmstylethree}%

\begin{document}

\title[ML-KFHE]{ML-KFHE: Multi-label ensemble classification algorithm exploiting sensor fusion properties of the Kalman filter}


\author*[1,2]{\fnm{Arjun} \sur{Pakrashi}}\email{arjun.pakrashi@ucd.ie}

\author[1,2]{\fnm{Brian} \sur{Mac\ Namee}}\email{brian.macnamee@ucd.ie}

\affil[1]{\orgdiv{School of Computer Science}, \orgname{University College Dublin}, \orgaddress{\city{Dublin}, \country{Ireland}}}

\affil[2]{\orgname{Insight Centre for Data Analytics}, \orgaddress{\city{Dublin}, \country{Ireland}}}


\abstract{Despite the success of ensemble classification methods in multi-class classification problems, ensemble methods based on approaches other than bagging have not been widely explored for multi-label classification problems. The Kalman Filter-based Heuristic Ensemble (KFHE) is an ensemble method that exploits the sensor fusion properties of the Kalman filter to combine several classifier models, and that has been shown to be very effective. This work proposes a multi-label version of KFHE, ML-KFHE, demonstrating the effectiveness of the KFHE method on multi-label datasets. Two variants are introduced based on the underlying component classifier algorithm, ML-KFHE-HOMER, and ML-KFHE-CC which uses HOMER and Classifier Chain (CC) as the underlying multi-label algorithms respectively. ML-KFHE-HOMER and ML-KFHE-CC sequentially train multiple HOMER and CC multi-label classifiers and aggregate their outputs using the sensor fusion properties of the Kalman filter. Extensive experiments and detailed analysis were performed on thirteen multi-label datasets and eight other algorithms, which included state-of-the-art ensemble methods. The results show, for both versions, the ML-KFHE framework improves the predictive performance significantly with respect to bagged combinations of HOMER (named E-HOMER), also introduced in this paper, and bagged combination of CC, Ensemble Classifier Chains (ECC), thus demonstrating the effectiveness of ML-KFHE. Also, the ML-KFHE-HOMER variant was found to perform consistently and significantly better than the compared multi-label methods including existing approaches based on ensembles.}




\maketitle

\section{Introduction}

A multi-class classification task assigns an object to at most one class.
Real-world classification problems exist, however, where an object can be
assigned to more than one class simultaneously \citep{DBLP:books/sp/HerreraCRJ16}.
In other words, an object can be \emph{labelled} with more than one
class at the same time. For example, an image of a landscape
may contain mountains, sea and sky and therefore it can be a member
of each of the corresponding classes \citep{BOUTELL20041757}. Similarly, music can be tagged with more than one
genre. Such problems are known as \emph{multi-label} classification
problems.

Formally, multi-label classification problems can be defined as follows.
Let $\boldsymbol{x}_{i}$ be a datapoint from a $d$-dimensional input
space $\mathcal{X}$ of real and/or categorical attributes. Also,
let the set of all possible labels for a specific multi-label classification
problem be $\mathcal{L}=\{\lambda_{1},\lambda_{2},\ldots,\lambda_{q}\}$,
from which a subset of labels, $\mathcal{L}_{i}\subseteq\mathcal{L}$,
is applicable to the datapoint $\boldsymbol{x}_{i}$. Here labels in $\mathcal{L}_{i}$
are called the \emph{relevant} labels, and $(\mathcal{L}-\mathcal{L}_{i})$
are called the \emph{irrelevant} labels for $\boldsymbol{x}_{i}$. Then
a typical multi-label dataset is defined as $\mathcal{D}=\{(\boldsymbol{x}_{i},\boldsymbol{y}_{i})\|1\le i\le n\}$,
where $n$ is the number of datapoints in the dataset; $\boldsymbol{x}_{i}=\{x_{i1,}x_{i2,}\ldots,x_{id}\}$
is a $d$-dimensional vector indicating the $i^{th}$ datapoint; and $\boldsymbol{y}_{i}=\{y_{i1,}y_{i2,}\ldots,y_{iq}\}$
is a binary vector indicating the label assignments $\mathcal{L}_{i}$
for the $i^{th}$ datapoint. Here $y_{ij}=1$ if $\lambda_{j}\in\mathcal{L}_{i}$,
that is, the $j^{th}$ label is relevant to the $i^{th}$ datapoint,
and $y_{ij}=0$ if $\lambda_{j}\notin\mathcal{L}_{i}$\emph{.}The
objective of multi-label classification is to learn a model $\boldsymbol{h}$,
which predicts the relevance of every label for a new datapoint $\boldsymbol{d}$, written as
$\boldsymbol{h}(\boldsymbol{d})$.

Multi-label classification algorithms can be categorised as either \emph{problem transformation} or \emph{algorithm adaptation} methods \citep{Tsoumakas07multi-labelclassification:}. Problem transformation methods---for example \emph{classifier chains} \citep{Read2011}---break the multi-label problem down into smaller multi-class classification problems. Algorithm adaptation methods---for example BPMLL \citep{Zhang:2006:MNN:1159162.1159294}---modify multi-class algorithms to directly train on multi-label datasets.

Ensemble classification methods train multiple component classifiers and aggregate them. Generally, ensemble methods perform better than the single component classifiers \citep{kelleher2015fundamentals} and ensemble classifiers based on boosting generally perform better than bagging methods \citep{narassiguin2016extensive}. In the multi-label classification literature, several methods have been proposed that combine multiple multi-label models to form an ensemble. These, however, are mostly problem transformation methods based on \emph{bagging} or majority voting based approaches to building ensembles \citep{Read2011,tremlc,tenenboim2010identification,Kocev2007,eps,rakel,rokach2014ensemble}. There are very few boosting approaches in the multi-label classification literature. AdaBoost.MH \citep{Schapire2000} is probably the most prominent boosting approach in multi-label classification, but does not perform well when compared to other methods \citep{MOYANO201833}. Although based on bagging and majority voting methods, algorithms like Classifier Chains (CC) \citep{Read2011} and RAkEL \citep{rakel} which still stay a very competitive \citep{MADJAROV20123084,pakrashi2016benchmarking} and has received test of time award in ECML 2017 and 2019, respectively. As boosting based methods in multi-class classification perform so much better than approaches based on bagging, but in the multi-label context they didn't perform as much as they perform in the multi-class domain, probably due to the high degree of imbalance which multi-label datasets have. This motivates the development of better ensembles for multi-label classification that are based on ideas similar to boosting.

\emph{Kalman Filter-based Heuristic Ensemble} (KFHE) \citep{PAKRASHI2019456} is a recently proposed algorithm that frames ensemble training as a state estimation problem which is solved using a Kalman filter \citep{kalman1960,maybeck1982stochastic}. Although Kalman filters are most commonly used to solve problems associated with time series data, this is not the case for KFHE. Rather, this work exploits the data fusion property of the Kalman filter to combine individual multi-class component classifier models to construct an ensemble. This can be interpreted as effectively being in between boosting and bagging, and therefore is expected to benefit from exploiting the advantages of both types of ensembles. Given the nature of this method, as it effectively falls between boosting and bagging, this can be especially helpful as multi-label datasets are inherently highly imbalanced and can be helpful to balance the drawbacks and benefits of both boosting and bagging.

By utilising the sensor fusion properties of the Kalman filter in the KFHE framework, this article proposes a multi-label classification algorithm ML-KFHE, and demonstrates its effectiveness using two variants of multi-label classification methods, \emph{ML-KFHE-HOMER} and \emph{ML-KFHE-CC}. Here, \emph{ML-KFHE-HOMER} and \emph{ML-KFHE-CC} which are ensembles of HOMER \citep{tsoumakas2008effective} and \emph{classifier chains} (CC) \citep{Read2011} algorithms, respectively, using the ML-KFHE framework.

To demonstrate the effectiveness of the ML-KFHE ensemble combination method, ensemble versions of CC and HOMER were compared. Ensemble version of CC called \emph{Ensemble Classifier Chains} (ECC) was proposed in \citep{Read2011}. Although bagged HOMER was not found in the literature, a simple bagged ensemble version of HOMER, named E-HOMER, was also proposed in this work to be able to directly compare the effectiveness of the ML-KFHE-HOMER. The source code of the algorithms are made online: \url{https://github.com/phoxis/kfhe-homer}

The contributions of this paper are:

\begin{itemize}
\item An ensemble multi-label classification method, ML-KFHE, which exploits the sensor fusion properties of a Kalman filter in KFHE algorithm to combine component classifiers. Two variants are presented, ML-KFHE-HOMER and ML-KFHE-CC, which incorporates HOMER and CC, respectively, as component classifiers in the KFHE framework.
\item An extensive experiment demonstrating the effectiveness of using the ML-KFHE method to combine multi-label classifiers over using bagging based multi-label ensemble algorithms, as well as demonstrating the overall effectiveness of the proposed methods.
\item Introduction of E-HOMER, a simple bagged ensemble version of HOMER to compare the ML-KFHE ensembling and bagged HOMER.
\end{itemize}

The remainder of the paper is structured as follows. Section \ref{sec:Literature-Review} describes existing multi-label ensemble classification algorithms, as well as the relevant aspects of CC and HOMER. Section \ref{sec:proposed} introduces the proposed ML-KFHE-HOMER and ML-KFHE-CC methods, by first introducing KFHE and then ML-KFHE. The design of the evaluation experiments performed is described in Section \ref{sec:Experiment}, and the results of these experiments are discussed in detail in Section \ref{sec:Results}. Finally, Section \ref{sec:Conclusion-and-Future} concludes the article and discusses directions for future work.

\section{Related Work}\label{sec:Literature-Review}

This section discusses the current state-of-the-art ensemble methods used for multi-label classification and then mentions the relevant aspects of the HOMER and CC algorithms, which are later used in this work.

\subsection{Ensemble Methods for Multi-label Classification}

In multi-label classification the same principles are used as in multi-class ensemble algorithms---although most of the multi-label ensembles are bagging based (in which multiple independent classification models vote on the relevance of each label). Some important multi-label ensemble classification algorithms will be described briefly next.

\emph{Ensemble Binary Relevance} (EBR) and ECC \citep{Read2011} are simple bagged version of the Binary Relevance and Classifier Chains classifier algorithms. Label Powerset models can also be bagged to form \emph{Ensemble Label Powerset} (ELP) models \citep{MOYANO201833}. \emph{Ensemble of Pruned Sets} (EPS) \citep{read2008multi} is also a simple bagged version of Pruned Set and prevents overfitting properties of the Pruned Set algorithm. RAkEL-o \citep{rakel} is also ensemble classifier algorithms, but instead of bagging it divides the label space into smaller overlapping subsets, then learn LP models for each of these subsets. RF-PCT \citep{Kocev2007}, is a Random Forest \citep{ROKACH2016111} of Probabilistic Clustering Trees (PCT) \citep{Blockeel:1998:TIC:645527.657456} and therefore is also a bagging type ensemble model. \emph{Triple Random Ensemble for Multi-label Classification (TREMLC)} \citep{tremlc} randomly performs several samples of datapoints and features at the same time and trains multiple LP models on each such sample combining the prediction from these trained models. \emph{Clustering Based for Multi-label Classification (CBMLC)} \citep{cbmlc} first performs a clustering on the feature space of the dataset to generate $k$ clusters, each of which ideally holds similar datapoints. Next, it trains any multi-label algorithm for each of the $k$ clusters. In summary, these bagged classifier algorithms range from combining existing component classifiers by training on bootstrap samples or sub-samples of the datapoints (EBR, ELP, EPS), then training different sub-samples of the datapoints along with different random subsets of the labels (RAkEL-o) or different random order of the labels (ECC). The other types performs bagging like subsampling but also considers subsampling or pre-processing the attributes as well (RF-PCT, TREMLC, CBMLC).



In boosting \emph{AdaBoost.MH} and \emph{AdaBoost.MR} \citep{Schapire2000} are the most frequently used multi-label ensemble classification models in existence. These two extensions of \emph{AdaBoost} \citep{freund1995desicion} for multi-label classification. AdaBoost.MH minimises the Hamming loss, and considers the labels independently and applies AdaBoost. Whereas, AdaBoost.MR considers pairwise label ranking by minimising a function similar to rankloss which penalises incorrect label ordering. Then it gives more weight to the datapoints which has a higher rank loss.

Some other less known algorithms are \emph{AdaBoost.MH$^{kr}$} \citep{sebastiani2000improved} which uses a relatively more complex model per boosting iteration by replacing the usual weak hypotheses with a sub-committee of weak hypotheses. \emph{RFBoost} \citep{ALSALEMI2016104} improves AdaBoost.MH by training the weak learner on only a few top ranked features instead of using all the features. RFBoost was experimented using two different feature ranking methods and shown to perform similar or slightly better than AdaBoost.MH.

\subsection{HOMER}


\emph{Hierarchy of Multi-label Classifiers} (\emph{HOMER}) \citep{tsoumakas2008effective}
also divides the multi-label dataset into smaller subsets of labels,
but in a hierarchical manner. This method divides the dataset based
on the labels, but establishes a hierarchical relationship between
the partitions. The root of the
hierarchy has all labels $\mathcal{L}$ and the entire dataset
associated with it. Every leaf node has one associated label $\lambda_{k}$.
Any internal node $v$ have only a subset of labels $\mathcal{L}_{v}\subset\mathcal{L}$
associated with itself, which is a union of the label subsets associated
to its children. Therefore, $\mathcal{L}_{v}=\cup\mathcal{L}_{c},\,c\in children(v)$,
where $children(v)$ indicates the children of the node $v$ in the
hierarchy. Each node only keeps the datapoints that have at least one
of the associated labels of the node in their relevant set. Therefore, the dataset
at the node $v$ is $\mathcal{D}_{v}=\{(\boldsymbol{x}_{j},\boldsymbol{y}_{j})\|\forall(\boldsymbol{x}_{j},\boldsymbol{y}_{j})\in parent(v),\exists_{k}\lambda_{k}\in\mathcal{L}_{v}\wedge y_{jk}=1\}$.
Each internal node $v$ is also associated with a meta label $\mu_{v}$,
which is associated with all the datapoints in the corresponding node,
and each internal node's datapoints are also associated with meta
labels assigned by its children. The meta labels are generated as
$D'_{v}=\{(\boldsymbol{x}_{j},\boldsymbol{y}'_{j})\|\boldsymbol{y}'_{j}=\cup_{c\in children(v)}\mu_{c}\}$.

The root node and all the internal nodes $v$ will have a trained
model $h_{v}$ associated to them, which is trained using the dataset
$D'_{v}$ with the target being the meta labels associated with each
datapoint in the child nodes. Utility of the meta labels is to
indicate which branch or branches in the hierarchy have to be followed
to predict the labels. A new datapoint $\boldsymbol{t}$ starts from
the root of the hierarchy and travels one or more paths from root to leaf.
All the labels represented by the leaves which are encountered by
the datapoint $\boldsymbol{t}$ are taken as the prediction.



\subsection{Classifier Chains}

\emph{Classifier Chains} \citep{Read2011} take a similar approach
to binary relevance but explicitly take the associations between labels
into account. Again a one-vs-all classifier is built for each label,
but these classifiers are chained together such that the outputs of
classifiers early in the chain (the relevance of specific labels)
are used as inputs into subsequent classifiers. Let $\tau:\{1\ldots q\}\rightarrow\{1\ldots q\}$
be a permutation function which gives a new ordering or a chain of
the labels $\lambda_{\tau(1)}\succ\lambda_{\tau(2)}\succ\cdots\succ\lambda_{\tau(q)}$.
For a label $\lambda_{\tau(l)}$ a dataset is formed $\mathcal{D}_{\tau(l)}=\{([\boldsymbol{x}_{i},pre_{\tau(j)}^{i}],y_{i\tau(l)})\|1\le i\le n\}$,
where $pre_{\tau(l)}^{i}=[y_{i\tau(1)},y_{i\tau(2)},\ldots,y_{i\tau(l-1)}]$,
is the concatenation of labels from the first label in the ordering
up to the previous label for the $i^{th}$ data point. Therefore, each
dataset, $\mathcal{D}_{\tau(l)}$, includes the original input space,
as well as the label space from $\lambda_{\tau(1)}$ up to $\lambda_{\tau(l-1)}$,
with a target label of $\lambda_{\tau(l)}$. This explicitly
imposes the dependency of the labels earlier in the chain on the labels
later in the chain. Next, for each $\mathcal{\mathcal{D}}_{\tau(l)}$
a binary classifier $h_{\tau(l)}$ is learned.

For prediction, the chain order generated using the permutation function
$\tau$ is followed. Prediction for $\boldsymbol{t}$ starts from first
predicting the $\lambda_{\tau(1)}$, $\hat{y}_{\tau(1)}=h_{\tau(1)}$,
then this label prediction is concatenated with $\boldsymbol{t}$ and
that is fed into next learned model in the chain. To predict $\lambda_{\tau(l)}$
all the labels $\lambda_{\tau(1)}$ to $\lambda_{\tau(l-1)}$ have
to be predicted in the chain order first allowing their predictions to be
concatenated. Therefore, predicting all the labels is done as follows.
$\hat{\boldsymbol{y}}=\{\hat{y}_{\tau(l)}\|th(h_{\tau(l)}([\boldsymbol{t},\hat{y}_{\tau(1)},\ldots,\hat{y}_{\tau(l-1)}])),1\le l\le q\}$.

\section{Proposed Method: ML-KFHE\label{sec:proposed}}

In this section first the relevant parts of KFHE will be mentioned then ML-KFHE will be described.

\subsection{KFHE Algorithm\label{subsec:KFHE}}

The discrete Kalman filter is a mathematical framework to estimate an unobservable state of a linear stochastic discrete time controlled process through noisy measurements \citep{faragher2012understanding}.

The \emph{Kalman filter-based Heuristic Ensemble} (KFHE) \citep{PAKRASHI2019456} is a multi-class ensemble algorithm which, unlike existing boosting or bagging methods, considers the ensemble to be trained as a hypothesis to be estimated within a hypothesis space. This approach considers the trained classifiers in an ensemble to be noisy measurements which it combines using a Kalman filter. In effect, KFHE behaves like a combination of both boosting and bagging. The remainder of this section describes how a Kalman filter can be used for static state estimation, before describing details of the KFHE approach.

%
Let there be a state, $y$, of a linear stochastic system has to be estimated, where $y$ cannot be observed directly. The state of the system can be estimated in two ways. Firstly, given an estimate of the state $\hat{y}_{t-1}$ with a related variance $p_{t-1}$ at time step $(t-1)$, a linear model is used to make an \emph{a priori} state estimate $\hat{y}_{t}^{-}$. The variance related to $\hat{y}_{t}^{-}$ is also updated to $p_{t}^{-}$. This variance can be imagined as the uncertainty of state. This is known as the \emph{time update} step. Secondly, an external sensor can be used to get an estimate through a \emph{measurement}, $z_{t}$, of the state with a related variance $r_{t}$, which can also be seen as the uncertainty of the measurement.
Given these two noisy state estimates, the \emph{a priori} estimate, $\hat{y}_{t}^{-}$, its related variance $p_{t}^{-}$, and the \emph{measurement} $z_{t}$, its related variance $r_{t}$, the Kalman filter combines them optimally to get an \emph{a posteriori} state estimate, $\hat{y}_{t}$, which potentially has a lower uncertainty than the previous two. This is known as the \emph{measurement update} step. The Kalman filter iterates through the time update and the measurement update steps. At iteration $t$, the \emph{a priori} estimate is used in the measurement update step to get an \emph{a posteriori} estimate, which is fed back to the time update in the next iteration as the \emph{a priori} estimate.

If the state to be estimated is assumed to be static, then the time update step is considered to be non-existent. This kind of scenario can occur in cases when, say, the voltage level of a DC battery or the altitude of a cruising aircraft is being estimated. In both of the cases, the DC voltage and the altitude of the aircraft is supposed to be constant, but unknown. In such cases, the measurement of the static state from a noisy sensor is repeatedly combined using the measurement update step.

The basic idea of KFHE is to view the ideal hypothesis for a specific classification problem as a static state to be estimated in a \emph{hypothesis space} \citep{10.1007/3-540-45014-9_1}. As in the above description, when estimating the static state, after $T$ iterations the estimate of the Kalman filter is essentially the combination or an \emph{ensemble} of a sensor output. Similarly, the component classifiers are combined in KFHE using the above principle. The equations used for the algorithm is as follows

\begin{equation}\label{eq:measurement_update}
    \hat{y}_{t} = \hat{\boldsymbol{y}}_{t-1} + k_{t}(z_{t}-\hat{\boldsymbol{y}}_{t-1})
\end{equation}

\begin{equation}\label{eq:k_update}
    k_{t} = p_{t-1} / (p_{t-1} + r_{t})
\end{equation}

\begin{equation}\label{eq:p_update}
    p_{t} = (1 - k_{t})p_{t-1}
\end{equation}

Here $z_{t}$, the measurement, can be an external source or sensor (voltage or altitude sensor), $r_{t}$ is the related measurement variance indicating the uncertainty of the estimate. The $k_{t}$ is the \emph{Kalman gain}, which optimally combines the \emph{a priori} estimate and the measurement. A complete and detailed explanation of Kalman filters can be found in \citep{faragher2012understanding,welch1995introduction}.


\begin{figure*}
\centering
\includegraphics[width=1.0\textwidth]{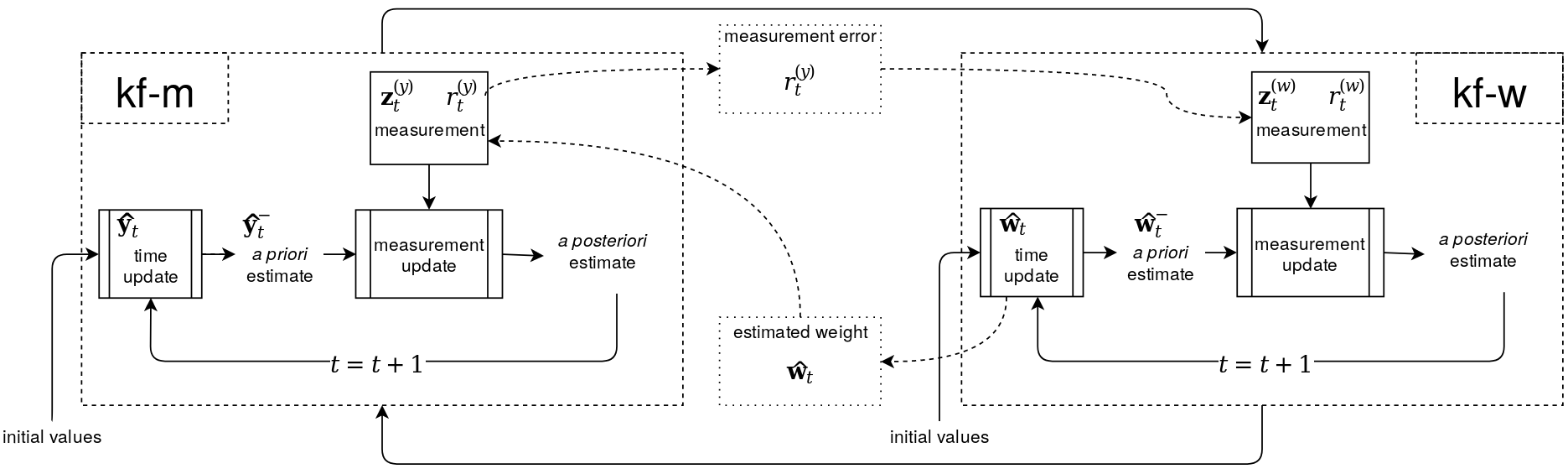}
\caption{The high level interactions between \emph{kf-m} and \emph{kf-w} \citep{PAKRASHI2019456} \label{fig:Overall-dataflow-between}}
\end{figure*}



KFHE has two components which interact with each other. The Kalman filter which estimates the ideal hypothesis (as described above) is called the model Kalman filter, abbreviated as \emph{kf-m}, estimates the final model or hypothesis by combining multiple noisy measurements. The other component named the weight Kalman filter, \emph{kf-w}, computes the weight using which the \emph{kf-m} performs the sampling of training datapoints for the component classifier.

The measurement in this case is defined as,

\begin{equation}\label{eq:measurement}
\boldsymbol{z}_{t}^{(y)} = (h_{t}(\mathcal{D}) + \boldsymbol{\hat{y}}_{t-1}) / 2
\end{equation}

Where $h_{t}=\mathcal{H}(\mathcal{D},\boldsymbol{\hat{w}}_{t-1})$ is a classifier model trained using algorithm $\mathcal{H}$ (decision tree, SVM, etc.) using a dataset defined by a set of datapoint weights updated in the previous iteration, $\boldsymbol{\hat{w}}_{t-1}$. A datapoint is weighted more if it was misclassified previously, and less if correctly classified which is similar to the approach taken in boosting. Although, unlike AdaBoost \citep{hastie2009multi}, the weights for the datapoints are determined by another Kalman filter, the weight Kalman filter or \emph{kf-w}.


\begin{table}[t!]
    \centering

    \caption{Intermediate representation of a state for KFHE and ML-KFHE. A trained model is represented using the prediction scores for the classes ($c_{1}$, $c_{2}$, and $c_{3}$) of a given set of datapoints. This representation is used with $\boldsymbol{\hat{y}}_{t}$, $\boldsymbol{z}_{t}$ and $h_{t}(\mathcal{D})$.}
    \label{tab:sr}
    \setlength{\tabcolsep}{10pt}
    \begin{tabular}{cccc}
        \hline
         & $c_{1}$ & $c_{2}$ & $c_{3}$\tabularnewline
        \hline
        $\boldsymbol{x}_{1}$ & $0.10$ & $0.89$ & $0.01$\tabularnewline
        $\boldsymbol{x}_{2}$ & $0.08$ & $0.27$ & $0.65$\tabularnewline
        $\vdots$ & $\vdots$ & $\vdots$ & $\vdots$\tabularnewline
        $\boldsymbol{x}_{n}$ & $0.77$ & $0.20$ & $0.03$\tabularnewline
        \hline
    \end{tabular}
\end{table}

Note that, the ensemble model $h_{t}$ cannot directly be used with the equations in Eq. \eqref{eq:measurement_update} and \eqref{eq:measurement}, therefore an intermediate proxy representation is used for the states in \emph{kf-m}. The intermediate representation of a trained model is the label-wise prediction scores of a given dataset by the model of the corresponding state, as shown in Table  \ref{tab:sr}. Therefore, the intermediate representation of a model (individual or ensemble) would be the prediction $\hat{\boldsymbol{y}}_{t}$ as shown in Table \ref{tab:sr}. For example, the first datapoint has the highest prediction score assigned to class-label $c_{2}$, and thus the first datapoint is considered as a member of class $c_{2}$ (among two other potential classes $c_{1}$ and $c_{3}$). This representation of a model is used as the state in the Kalman filter framework. In the final estimated state the class assignment is done by taking the class with the highest score.


The \emph{kf-w} estimates $\boldsymbol{\hat{w}}_{t}$, which is a vector of weights to be used by the measurement step of \emph{kf-m}. \emph{kf-w} is otherwise identical to \emph{kf-m}.




The training step stores the component classifiers $h_{t}$ and the Kalman gains $k_{t}^{(y)}$. When a new datapoint is encountered during the prediction step the Eq. (\ref{eq:measurement_update}) is repeatedly used using the component classifiers $h_{t}$ and the Kalman gains $k_{t}^{(y)}$ found during the training stage.

An overall interaction of the \emph{kf-m} and \emph{kf-w} is shown in Figure \ref{fig:Overall-dataflow-between}. The superscript $(y)$ indicates that the variables are related to \emph{kf-m}, and the superscript $(w)$ indicates these variables are related to \emph{kf-w}. $\boldsymbol{\hat{y}}$ is the state estimate by \emph{kf-m}, and $\boldsymbol{\hat{w}}_{t}$ is estimated by \emph{kf-w}.

The setting of the measurements and the related errors are the heuristic components of the method, which are set by making assumptions. A detailed explanation of the concept, derivation and explanation of KFHE can be found in \citep{PAKRASHI2019456}, and a few related work utilising the Kalman filter based framework to combine classifier models can be found in \citep{pakrashi2020kalmantune,okfse}.

\subsection{ML-KFHE\label{sec:KFHE-HOMER}}

In this work, ML-KFHE, proposes a multi-label classification algorithm by combining multiple multi-label classifier models exploiting the sensor fusion properties of the Kalman filter. Depending on which underlying multi-label classifiers are ensembled, two variants are proposed in this work. ML-KFHE-HOMER, which ensembles multiple HOMER models and ML-KFHE-CC which ensembles multiple CC models. HOMER is a multi-label classification method that has been shown to have competitive performance with other leading approaches \citep{MADJAROV20123084,pakrashi2016benchmarking}. HOMER was selected for this task because it has a lower training time which makes it suitable for this purpose to train more ensemble components in a shorter time. Also, CC was chosen as the existing ECC \citep{Read2011} already attains very good classification performance \citep{MOYANO201833} as well as takes label associations into consideration.


As explained in Section \ref{subsec:KFHE}, there are two components of KFHE: \emph{kf-m} that estimates the hypothesis, and \emph{kf-w} that computes the weights of the training datapoints during each measurement. To make KFHE work in a multi-label setting, the measurements of the \emph{kf-m} and \emph{kf-w} steps were adapted in this work.


For ML-KFHE, the measurement at each step is the average of a trained multi-label classifier and the previous estimate of the ensemble as shown in Eq. (\ref{eq:measurement}). The related measurement uncertainty $r_{t}^{(y)}$ is the Hamming loss ($hloss$) \citep{zhang2014review} of the trained model. Each multi-label model at every step is trained on different weights, $\boldsymbol{\hat{w}}_{t}$, assigned to different datapoints, where the weights are determined by the \emph{kf-w} component. The \emph{kf-w} estimates one single vector of weights $\boldsymbol{\hat{w}}_{t}$ using which a sampling with replacement of the training dataset is done. Although the measurement $\boldsymbol{z}_{t}^{(w)}$ for \emph{kf-w} is taken as per-datapoint weighted Hamming loss, which can be defined as follows

\begin{equation}\label{eq:measurement_weight_kfhehomer}
\boldsymbol{z}_{t}^{(w)}=[z_{ti}^{(w)}\|z_{ti}^{(w)}=\hat{w}_{ti}\times exp (hloss (\boldsymbol{x}_{i}, {\boldsymbol{l}_{i}}))\,\,\,1\le i\le n]
\end{equation}

Eq. (\ref{eq:measurement_weight_kfhehomer}) uses Hamming loss and the exponential function (KFHE-e variant) to highlight misclassified datapoints in the measurement which will later be used   by the measurement update step to get the weights $\boldsymbol{\hat{w}}_{t}$ to be used in \emph{kf-m} in the next iteration. In this case the related uncertainty is calculated as in KFHE.

The model $h_{t}$ in this case is a trained multi-label classifier model $\mathcal{H}(\mathcal{D}, \boldsymbol{\hat{w}}_{t-1}, \mathcal{M})$. Here $\mathcal{H}$ is CC or HOMER (the underlying multi-label classifier) algorithm and $\mathcal{M}$ is the hyperparameters of $\mathcal{H}$. To weight the datapoints for training, the multi-label lassifiers are trained using samples using the distribution $\boldsymbol{\hat{w}}_{t-1}$, the last updated weights. Based on the underlying multi-label classification algorithm two variants are presented in this work.

\begin{itemize}
  \item \textbf{ML-KFHE-HOMER:} This variant uses HOMER as the underlying classifier. To train each component HOMER classifier the following three hyperparameter are modified. $\mathcal{M} = \{\mathcal{C},k,\phi\}$. Here $\mathcal{C}$ is the clustering algorithm used by HOMER is randomly selected from \{\emph{random}, \emph{k-means}, \emph{balanced k-means}\}, $k$ is the number of clusters which is randomly selected too. Also, the kernel $\phi$ of the underlying SVM used by HOMER, is also selected randomly.

  \item \textbf{ML-KFHE-CC:} This variant uses CC as the underlying classifier and the hyperparameters adjusted randomly to increase diversely of the models in this case are $\mathcal{M} = \{\mathcal{O},\phi\}$. Here, $\mathcal{O}$ is the chain order for a component CC classifier, which is selected randomly. Also, like before, the kernel type $\phi$ of underlying SVM used by CC is selected randomly.
\end{itemize}


Next, the measurement is done using Eq. (\ref{eq:measurement}).

The above method is applied to increase diversity of the models. The reason to increase diversity, for example, the HOMER models in the case of ML-KFHE-HOMER by randomly selecting the clustering algorithm, cluster size, and the SVM kernel type is as follows. Given a set of different HOMER models trained using different hyperparameters, many of them may lead to a poor measurement. The ML-KFHE framework combines the measurements based on the measurement errors. If the measurement uncertainty $r_{t}^{(y)}$ is higher than the uncertainty of the ensemble found up to the $t$th iteration $p_{t}^{(y)}$, then the measurement is weighted less and the Kalman gain is lower than $0.5$, and when the measurement error is lower the measurement is incorporated more, as a result of the Kalman gain being greater than $0.5$. Therefore, based on this property, the HOMER models which have a poor performance will have a much less impact on the entire ensemble, whereas a more accurate HOMER model will have more impact on the entire ensemble. This also applies on the ML-KFHE-CC version, where the diversity is induced by selecting the random chain order and the randomly selected underlying SVM kernel.

The values of $\boldsymbol{\hat{y}}_{0}$, $p_{0}^{(y)}$ and $\boldsymbol{\hat{w}}_{0}$, $p_{0}^{(w)}$ have to be initialised. $\boldsymbol{\hat{y}}_{0}$ is initialised using a single $\mathcal{H}$ classifier model, $h_{0}$. The value of $p_{0}^{(y)}$ is set to $1$ indicating maximum uncertainty. Equal weight is given to every point in $\boldsymbol{\hat{w}}_{0}$, and $p_{0}^{(w)}$ is also initialised with $1$.

Algorithm \ref{alg:kfhe_homer_train} shows the ML-KFHE training algorithm . The superscripts $(y)$ and $(w)$ indicate that the corresponding variables are related to \emph{kf-m} and \emph{kf-w} respectively. On Lines 7-17 the different hyperparameters of HOMER are selected randomly. Next, the component classifier model is trained on Line 18, and the measurement is done on Line 19. Line 21 computes the Kalman gain, $k_{t}^{(y)}$, for \emph{kf-m} and Line 22 computes the proxy representation of the ensemble $\boldsymbol{\hat{y}}_t{}$, based on the ML-KFHE ensemble predictions on the training dataset. The \emph{kf-w} steps are similar and are performed on Lines 25-29f. The process runs until a maximum number of ensemble iterations $T$.

The prediction algorithm is the same as for KFHE and is shown in Algorithm \ref{alg:kfhe_homer_pred}. Here the trained models and the Kalman gain values learned during the training along with a new query datapoint is given. Using the models in Line 5 the Kalman gain is repeatedly used to combine the measurements on Line 4. After $T$ iterations the predicted labels for the new datapoint $\boldsymbol{d}$, the estimate $\boldsymbol{\hat{y}}_{T}^{(y)}$ are returned. To find the label assignments, these scores are thresholded at $0.5$.

\begin{algorithm}[H]
\caption{ML-KFHE training}\label{alg:kfhe_homer_train}
\begin{algorithmic}[1]
\Procedure{train}{$\mathcal{D}=\{(\boldsymbol{x}_{i}, \boldsymbol{l}_{i})\|1\le i\le n\}$, $T$}
\State $p_{0}^{(w)}=1$, $\boldsymbol{\hat{w}}_{0} = [1/n, \ldots,  1/n]$
\State $h_{t} = \mathcal{H}(\mathcal{D}, \boldsymbol{\hat{w}}_{0}, \mathcal{C}, k, \phi)$, $\boldsymbol{\hat{y}}_{0} = h_{0}(\mathcal{D})$
\State $t=1$
\For{$t \le T$}
\State \rule{0.6\columnwidth}{1px}\Comment{\textit{\emph{kf-m}}}
\If{$\mathcal{H}$ is HOMER}
\State $\mathcal{M} = \{\mathcal{C}, k, \phi \}$
\State Choose $\mathcal{C} \in \{\text{\emph{k-means}}, \text{\emph{balanced k-means}}, \text{\emph{random}}\}$ randomly
\State Choose $k \in \{2, \ldots, \ceil{\sqrt{\|\mathcal{L}\|}}\}$ randomly
\State Choose $\phi \in \{\text{\emph{linear}}, \text{\emph{radial}}\}$ randomly
\ElsIf{$\mathcal{H}$ is CC}
\State Select $\mathcal{M} = \{\mathcal{O}, \phi\}$
\State $\mathcal{O}  $ is a random label ordering
\State Choose $\phi \in \{\text{\emph{linear}}, \text{\emph{radial}}\}$ randomly
\EndIf
\State $h_{t} = \mathcal{H}(\mathcal{D}, \boldsymbol{\hat{w}}_{t-1}, \mathcal{M})$
\State $\boldsymbol{z}_{t}^{(y)} = (h_{t}(\mathcal{D}) + \boldsymbol{\hat{y}}_{t-1})/2$\Comment{Measurement}
\State $r_{t}^{(y)}={\text{\emph{hloss}} (\mathcal{D}, \boldsymbol{z}_{t}^{(y)}) }$
\State $k^{(y)}_{t}=p_{t-1}^{(y)}/(p_{t-1}^{(y)}+r_{t}^{(y)})$\Comment{Kalman gain}
\State $\boldsymbol{\hat{y}}_{t}=\boldsymbol{\hat{y}}_{t-1}+k_{t}^{(y)}(\boldsymbol{z}_{t}^{(y)}-\boldsymbol{\hat{y}}_{t-1})$\Comment{Measurement update}

\State $p_{t}^{(y)}=(1-k_{t}^{(y)})p_{t-1}^{(y)}$

\State \rule{0.6\columnwidth}{1px}\Comment{\textit{\emph{kf-w}}}
\State $\boldsymbol{z}_{t}^{(w)}=[z_{ti}^{(w)}\|z_{ti}^{(w)}=\hat{w}_{ti}\times exp (hloss (\boldsymbol{x}_{i}, {\boldsymbol{l}_{i}}))\,\,\,1\le i\le n]$
\State $r_{t}^{(w)}={r_{t}^{(m)}}$
\State $k^{(w)}_{t}=p_{t-1}^{(w)}/(p_{t-1}^{(w)}+r_{t}^{(w)})$\Comment{Kalman gain}
\State $\boldsymbol{\hat{w}}_{t}=\boldsymbol{\hat{w}}_{t-1}+k_{t}^{(w)}(\boldsymbol{z}_{t}^{(w)}-\boldsymbol{\hat{w}}_{t-1})$\Comment{Measurement update}
\State $p_{t}^{(w)}=(1-k_{t}^{(w)})p_{t-1}^{(w)}$
\State \rule{0.6\columnwidth}{1px}
\State $t = t + 1$

\EndFor
\State \textbf{return} $(\{h_{t}, k_{t}^{(y)}\|\forall_{1\le t \le T}\})$
\EndProcedure
\end{algorithmic}
\end{algorithm}

\begin{algorithm}[H]
\caption{ML-KFHE prediction}\label{alg:kfhe_homer_pred}
\begin{algorithmic}[1]
\Procedure{predict}{$\boldsymbol{d},\{h_{t}, k_{t}^{(y)}\|\forall_{1\le t \le T}\}, T$}
\State $\boldsymbol{\hat{y}}_{0}^{(y)}=h_{0}(\boldsymbol{x})$, $t=1$
\For{$t \le T$}
  \State $\boldsymbol{\hat{z}}_{t}^{(y)} = (h_{t}(\boldsymbol{d}) + \boldsymbol{\hat{y}}_{t-1})/2$\Comment{Measurement}
  \State $\boldsymbol{\hat{y}}_{t}=\boldsymbol{\hat{y}}_{t-1}+k_{t}^{(y)}(\boldsymbol{z}_{t}^{(y)}-\boldsymbol{\hat{y}}_{t-1})$\Comment{Measurement update}
  \State $t = t + 1$
\EndFor
\State \textbf{return} ($\boldsymbol{\hat{y}}_{T}^{(y)}$)
\EndProcedure
\end{algorithmic}
\end{algorithm}
\begin{itemize}
\item \textbf{E-HOMER:} A simple bagged version of HOMER, E-HOMER, is also introduced in this section mainly with the intension to compare with ML-KFHE-HOMER to evaluate the effectiveness of ML-KFHE-HOMER. The hyperparameters for each HOMER model in the ensemble (the cluster type, the number of clusters, and the type of underlying SVM kernel) are all selected randomly, as in ML-KFHE-HOMER. The difference between E-HOMER and ML-KFHE-HOMER is that the combination of ML-KFHE-HOMER uses the ML-KFHE framework to ensemble the HOMER component classifier models and E-HOMER ensembles the component HOMER classifiers using simple bagging. Algorithm \ref{alg:e_homer_train} describes the E-HOMER training process, and Algorithm \ref{alg:e_homer_pred} describes the prediction process.
\end{itemize}

\begin{algorithm}[H]
\caption{E-HOMER training}\label{alg:e_homer_train}
\begin{algorithmic}[1]
\Procedure{train}{$\mathcal{D}=\{(\boldsymbol{x}_{i}, \boldsymbol{l}_{i})\|1\le i\le n\}$, $T$}
\State $h_{t} = \mathcal{H}(\mathcal{D}, \boldsymbol{\hat{w}}_{0}, \mathcal{C}, k, \phi)$, $\boldsymbol{\hat{y}}_{0} = h_{0}(\mathcal{D})$
\State $t=1$
\For{$t \le T$}
\State Randomly select $\mathcal{C}$, $k$ and $\phi$, where\\
\ \ \ \ \ \ \ \ \ \ \ \ \ \ \ \ $C \in \{\text{\emph{k-means}}, \text{\emph{balanced k-means}}, \text{\emph{random}}\}$, \\
\ \ \ \ \ \ \ \ \ \ \ \ \ \ \ \ $k \in \{2, \ldots, \ceil{\sqrt{\|\mathcal{L}\|}}\}$, \\
\ \ \ \ \ \ \ \ \ \ \ \ \ \ \ \ $\phi \in \{\text{\emph{linear}}, \text{\emph{radial}}\}$
\State $\boldsymbol{b} = bootstrap\_sample (2\times n)$
\State $h_{t} = \mathcal{H}(\mathcal{D}, \boldsymbol{b}, \mathcal{C}, k, \phi)$ 
\State $t = t + 1$

\EndFor
\State \textbf{return} $(\{h_{t} \|\forall_{1\le t \le T}\})$
\EndProcedure
\end{algorithmic}
\end{algorithm}


\begin{algorithm}[H]
\caption{E-HOMER prediction}\label{alg:e_homer_pred}
\begin{algorithmic}[1]
\Procedure{predict}{$\boldsymbol{d},\{h_{t}\|\forall_{1\le t \le T}\}, T$}
\State $\boldsymbol{\hat{y}}_{0}^{(y)}=h_{0}(\boldsymbol{x})$, $t=1$
\For{$t \le T$}
  \State $\boldsymbol{\hat{y}}_{t}=\boldsymbol{\hat{y}}_{t-1} + h_{t}(\boldsymbol{d})$
  \State $t = t + 1$
\EndFor
\State \textbf{return} ($\frac{\boldsymbol{\hat{y}}_{T}^{(y)}}{T}$)
\EndProcedure
\end{algorithmic}
\end{algorithm}

\section{Experiment\label{sec:Experiment}}

To evaluate the effectiveness of ML-KFHE, experiments were performed on thirteen well-known multi-label benchmark datasets\footnote{Dataset sources: http://mulan.sourceforge.net/datasets-mlc.html, http://www.uco.es/kdis/mllresources/} listed in Table \ref{tab:datasets}. In Table \ref{tab:datasets}, different properties of the multi-label datasets are summarised. \emph{Instances}, \emph{Inputs} and \emph{Labels} are the number of datapoints, the dimension of the datapoints, and the number of labels, respectively. \emph{Labelsets} indicates the number of unique combinations of labels. \emph{Cardinality} measures the average number of labels assigned to each datapoint and \emph{MeanIR} \citep{scumble} indicates the degree of imbalance of the labels, where higher values indicate higher imbalance.

\begin{table}[h!]
\caption{Multi-label datasets used in this work\label{tab:datasets}}
\centering
\resizebox{\columnwidth}{!}{ %
\centering\centering
\setlength{\tabcolsep}{3pt}
\begin{tabular}{lrrrHrrr}
\hline
Dataset          & Instances  & Inputs  & Labels  & Labelsets  & Labelsets  & Cardinality & MeanIR \tabularnewline
\hline
flags            & 194        & 26      & 7       & 54         & 24         & 3.392       & 2.255  \tabularnewline
yeast            & 2417       & 103     & 14      & 198        & 77         & 4.237       & 7.197  \tabularnewline
scene            & 2407       & 294     & 6       & 15         & 3          & 1.074       & 1.254  \tabularnewline
emotions         & 593        & 72      & 6       & 27         & 4          & 1.869       & 1.478  \tabularnewline
medical          & 978        & 1449    & 45      & 94         & 33         & 1.245       & 89.501 \tabularnewline
enron            & 1702       & 1001    & 53      & 753        & 573        & 3.378       & 73.953 \tabularnewline
birds            & 322        & 260     & 20      & 89         & 55         & 1.503       & 13.004 \tabularnewline
genbase          & 662        & 1186    & 27      & 32         & 10         & 1.252       & 37.315 \tabularnewline
cal500           & 502        & 68      & 174     & 502        & 502        & 26.044      & 20.578 \tabularnewline
llog             & 1460       & 1004    & 75      & 304        & 189        & 1.180       & 39.267 \tabularnewline
foodtruck        & 407        & 21      & 12      & 74         & 116        & 2.290       & 7.094  \tabularnewline
Water\_quality   & 1060       & 16      & 14      & 692        & 852        & 5.073       & 1.767  \tabularnewline
PlantPseAAC      & 978        & 440     & 12      & 8          & 32         & 1.079       & 6.690  \tabularnewline
\hline
\end{tabular}}
\end{table}

A brief description of the datasets are as follows:

\begin{itemize}
  \item \emph{flags} \cite{goncalves2013genetic}: Different properties of flags are used to predict seven different colours of flags. Each colour is considered as a label.
  \item \emph{yeast} \cite{Elisseeff01akernel}: Is a widely used datasest from the biological domain where genes are associated one or more of 14 different biological functions (labels).
  \item \emph{scene} \cite{BOUTELL20041757}: Several pictures of the scene is processed picture of scenary. Each scene can have one or more of the following labels: "beach", "sunset", "field", "fall-foliage", "mountain" and "urban".
  \item \emph{emotions} \cite{trohidis2008multi}: Each datapoint represents a piece of music. Each instance can be labelled with six emotions: "sad-lonely", "angry-aggressive", "amazed-surprised", "relaxing-calm", "quiet-still" and "happy-pleased".
  \item \emph{medical} \cite{pestian_et_al}: Each datapoint represents a document which includes a brief free text summary of patient symptom history and their prognosis, with 45 ICD-9-CM (International Classification of Diseases, Ninth Revision, Clinical Modification) codes\footnote{https://www.cdc.gov/nchs/icd/icd9cm.htm}.
  \item \emph{enron} \cite{read2010scalable}: This dataset is a version of Enron email corpus \footnote{http://www.cs.cmu.edu/~enron/} found in \cite{read2010scalable}, where each leaf categories in the original Enron corpus hierarchy is considered as a label, making a total of 53 labels.
  \item \emph{birds} \cite{briggs_et_al}: Each of the datapoint is a bird song recording and other natural sound recordings which can be labelled with 19 bird species or none of them. For this thesis, an additional label is added indicating "not a bird song" is added to avoid empty prediction problem \cite{empty_pred_liu}.
  \item \emph{genbase} \cite{diplaris_et_al}: Similar to the \emph{yeast} dataset, this dataset is from the microbiology domain concerned with gene functions, where each gene can be labelled with 27 labels.
  \item \emph{cal500} \cite{turnbull_et_al}: Each datapoint is a western song, which were hand annotated with genre, emotions, etc. Then each of the song/datapoint is labelled with 174 possible "musically relevant" labels in \cite{turnbull_et_al} and named Computer Audition Lab 500 dataset.
  \item \emph{llog} \cite{read2010scalable}: This dataset is a version of data found in the Language Log Forum \footnote{http://languagelog.ldc.upenn.edu/nll/} used in \cite{read2010scalable}. This assigns 75 possible topics (labels) to each datapoint describing the document.
  \item \emph{foodtruck} \cite{rivolli2017food}: This dataset was created by using the answers provided by the 407 survey on 21 objective questions about food truck preferences of the participants and their users' profile. Food truck preference questions were ingredients, place to sit, menu, hygiene preferences, along with personal users' information questions were age group, average income etc. The responses were recorded as categorical attributes.
  \item \emph{Water\_quality} \cite{blockeel1999simultaneous}: This dataset is used to predict the quality of water of Slovenian rivers, using 16 attributes such as, the temperature, pH, hardness, nitrous oxide or carbon di-oxide.
  \item \emph{PlantPseAAC} \cite{xu2016multi}: This dataset contains 978 sequences for Plant species. Gene ontology, amino acids, pseudo-amino acids and diptide components are provided. There are 12 labels indicating subcellular locations (cell membrace, cell wall, chloroplast, cytoplasm, endoplasmic reticulum, extracellular, golgi apparatus, mitochondrion, nucleus, peroxisome, plastid, and vacuole), which needs to be predicted.
\end{itemize}

Label-based macro-averaged F-Score \citep{zhang2014review} was used to  measure the performance of models in these  evaluations. This was chosen over Hamming loss, which has been used in several previous studies (e.g. \citep{brknn,mlknn,iblrml}) because in the highly imbalanced multi-label datasets used in this study (see the high MeanIR scores for several datasets in Table \ref{tab:datasets}) with Hamming loss performance on the majority classes may overwhelm the performance of the minority class.

%
For all evaluations $2$ times $5$ fold cross-validation experiments were performed. While generating the folds for cross-validation method a multi-label specific stratification method, iterative stratification \citep{iterstat}, was used.

Performance of ML-KFHE-HOMER and ML-KFHE-CC was compared with various algorithms. The hyperparameter configurations of each will be described next.

ECC, RAkEL, state-of-the-art ensemble based multi-label classifiers, AdaBoost.MH, RF-PCT and E-HOMER (a bagging-based ensemble using HOMER, Section \ref{sec:KFHE-HOMER}). ECC and RAkEL was specifically compared as they were the top performer in the extensive experiment in \citep{MOYANO201833}. The hyperparameter setting for RAkEL (named RAkEL2) was used as in \citep{MOYANO201833} as it is shown to perform very well. RAkEL2 is RAkEL with label subset size of $3$ and number of such random label overlapping subsets is $2q$, where the $q$ is the number of labels in the corresponding dataset. Support vector machine (SVM) is used as the underlying classifiers for ECC, RAkEL and all algorithms.

The compared hyperparameter tuned individual classifier (non-ensemble) algorithms compared are as follows. For, CC, HOMER-K (using k-means clustering) and HOMER-B (using balanced clustering) models were included to understand how much the ensembles led to improved performance over single base models when they are hyperparameter tuned for performance for each dataeset. During all the experiments, the sample size of training data (bag fraction) is $2n$, or twice the size of original number of datapoints for all the ensemble algorithms. The cluster size for HOMER was selected using the best values found in the benchmark experiments in \citep{pakrashi2016benchmarking}. The HOMER and CC models used support vector machines (SVM) as their underlying learner (as all the others), as they have proved to perform very well \citep{MADJAROV20123084,pakrashi2016benchmarking,MOYANO201833}.

Now the configuration of the proposed methods will be explained. At each iteration of E-HOMER and ML-KFHE-HOMER, for the underlying HOMER algorithm, the type of clustering, $\mathcal{C}$, was selected randomly from \{\emph{balanced k-means}, \emph{k-means}, \emph{random}\}, the number of clusters $k$ was selected randomly from the range $k \in \{2,\ldots,\ceil{\sqrt{\|\mathcal{L}\|}}\}$, and the SVM kernel $\phi$ from $\{linear, radial\}$, at each ensemble iteration. For ML-KFHE-CC, the kernel types $\phi$ for each of the base SVM models was selected at each ensemble iteration randomly, from $\{linear, radial\}$. The chain ordering of each component CC classifier was also selected randomly at each ensemble iteration.

For ECC and E-HOMER the bootstrap sample was selected to be twice the size of the training dataset, to keep it consistent with the ML-KFHE variants. A total of $100$ component classifiers were trained for all the ogher ensemble algorithms. Therefore, the experimental environment were kept identical for all ensemble methods for a fair comparison.

ML-KFHE and E-HOMER are implemented in R scripting language\footnote{A version of ML-KFHE and E-HOMER is available at:\\ https://github.com/phoxis/kfhe-homer}, AdaBoost.MH and RF-PCT was used from the MULAN \citep{mulan} and CLUS \footnote{https://dtai-static.cs.kuleuven.be/clus/} libraries respectively, and for the other methods the \emph{utiml} library \citep{rivolli2018utiml} is used.

\section{Results\label{sec:Results}}

\begin{landscape}

\begin{table*}[!t]
\caption{Experiment results. Values in cells are mean label-based macro-averaged F-Scores from the cross-validation experiments, and their standard deviations. The rank of each score for a dataset across the algorithms compared is shown in parenthesis. The last row shows the average
rank of each algorithms.\label{tab:big_table}}
\centering
\setlength{\tabcolsep}{4pt}
\resizebox{1.0\columnwidth}{!}{%
\begin{tabular}{rllllllllll}
  \hline
                  & ML-KFHE-HOMER            & E-HOMER                  & ML-KFHE-CC                & ECC                       & HOMER-B                  & CC                       & RAkEL2                   & HOMER-K                  & RF-PCT                   & AdaBoost.MH              \\
  \hline
  flags           & 0.6862 $\pm$ 0.02 (1)    & 0.6710 $\pm$ 0.05 (2)    & 0.6271 $\pm$ 0.04 (7\ \ ) & 0.6387 $\pm$ 0.04 (6\ \ ) & 0.6681 $\pm$ 0.04 (3\ )  & 0.5957 $\pm$ 0.05 (8)    & 0.5891 $\pm$ 0.05 (9)    & 0.6484 $\pm$ 0.03 (4\ )  & 0.6473 $\pm$ 0.05 (5\ )  & 0.5796 $\pm$ 0.08 (10)   \\
  yeast           & 0.4899 $\pm$ 0.01 (1)    & 0.4400 $\pm$ 0.04 (5)    & 0.4661 $\pm$ 0.01 (2\ \ ) & 0.4403 $\pm$ 0.02 (4\ \ ) & 0.3812 $\pm$ 0.02 (7\ )  & 0.4565 $\pm$ 0.01 (3)    & 0.4288 $\pm$ 0.01 (6)    & 0.3524 $\pm$ 0.02 (9\ )  & 0.3671 $\pm$ 0.01 (8\ )  & 0.1222 $\pm$ 0.00 (10)   \\
  scene           & 0.8090 $\pm$ 0.02 (1)    & 0.8008 $\pm$ 0.02 (2)    & 0.8005 $\pm$ 0.01 (3\ \ ) & 0.7806 $\pm$ 0.02 (6\ \ ) & 0.7777 $\pm$ 0.02 (7\ )  & 0.7818 $\pm$ 0.02 (5)    & 0.7990 $\pm$ 0.02 (4)    & 0.2011 $\pm$ 0.03 (9\ )  & 0.7161 $\pm$ 0.01 (8\ )  & 0.0000 $\pm$ 0.00 (10)   \\
  emotions        & 0.7046 $\pm$ 0.03 (1)    & 0.6891 $\pm$ 0.03 (4)    & 0.6731 $\pm$ 0.03 (6.5)   & 0.6731 $\pm$ 0.03 (6.5)   & 0.6964 $\pm$ 0.03 (2\ )  & 0.6595 $\pm$ 0.04 (9)    & 0.6915 $\pm$ 0.03 (3)    & 0.6809 $\pm$ 0.04 (5\ )  & 0.6617 $\pm$ 0.02 (8\ )  & 0.0573 $\pm$ 0.02 (10)   \\
  medical         & 0.6387 $\pm$ 0.02 (1)    & 0.6254 $\pm$ 0.03 (3)    & 0.6235 $\pm$ 0.02 (4\ \ ) & 0.6274 $\pm$ 0.02 (2\ \ ) & 0.5505 $\pm$ 0.03 (7\ )  & 0.6114 $\pm$ 0.02 (6)    & 0.6126 $\pm$ 0.03 (5)    & 0.3674 $\pm$ 0.04 (9\ )  & 0.3356 $\pm$ 0.05 (10)   & 0.4933 $\pm$ 0.03 (8\ )  \\
  enron           & 0.2612 $\pm$ 0.02 (1)    & 0.2587 $\pm$ 0.02 (2)    & 0.2458 $\pm$ 0.02 (3\ \ ) & 0.2435 $\pm$ 0.03 (4\ \ ) & 0.1888 $\pm$ 0.01 (7\ )  & 0.1985 $\pm$ 0.02 (6)    & 0.1877 $\pm$ 0.01 (8)    & 0.2123 $\pm$ 0.02 (5\ )  & 0.1760 $\pm$ 0.04 (9\ )  & 0.1490 $\pm$ 0.03 (10)   \\
  birds           & 0.3928 $\pm$ 0.05 (1)    & 0.3834 $\pm$ 0.04 (2)    & 0.3586 $\pm$ 0.04 (3\ \ ) & 0.3463 $\pm$ 0.06 (4\ \ ) & 0.3256 $\pm$ 0.05 (6\ )  & 0.3297 $\pm$ 0.04 (5)    & 0.3161 $\pm$ 0.05 (8)    & 0.3203 $\pm$ 0.03 (7\ )  & 0.2176 $\pm$ 0.04 (9\ )  & 0.1105 $\pm$ 0.08 (10)   \\
  genbase         & 0.9402 $\pm$ 0.03 (1)    & 0.8911 $\pm$ 0.03 (6)    & 0.9273 $\pm$ 0.03 (3\ \ ) & 0.9293 $\pm$ 0.02 (2\ \ ) & 0.7534 $\pm$ 0.06 (8\ )  & 0.9245 $\pm$ 0.02 (4)    & 0.9217 $\pm$ 0.03 (5)    & 0.8810 $\pm$ 0.03 (7\ )  & 0.2333 $\pm$ 0.08 (10)   & 0.2593 $\pm$ 0.04 (9\ )  \\
  cal500          & 0.1458 $\pm$ 0.01 (1)    & 0.1455 $\pm$ 0.01 (2)    & 0.1341 $\pm$ 0.01 (3\ \ ) & 0.1310 $\pm$ 0.01 (4\ \ ) & 0.0331 $\pm$ 0.01 (10)   & 0.0859 $\pm$ 0.01 (6)    & 0.0537 $\pm$ 0.00 (8)    & 0.0642 $\pm$ 0.02 (7\ )  & 0.1231 $\pm$ 0.03 (5\ )  & 0.0502 $\pm$ 0.01 (9\ )  \\
  llog            & 0.2315 $\pm$ 0.01 (5)    & 0.2308 $\pm$ 0.01 (6)    & 0.2267 $\pm$ 0.01 (7\ \ ) & 0.2236 $\pm$ 0.01 (9\ \ ) & 0.2253 $\pm$ 0.01 (8\ )  & 0.2454 $\pm$ 0.02 (2)    & 0.2404 $\pm$ 0.02 (4)    & 0.1683 $\pm$ 0.02 (10)   & 0.2460 $\pm$ 0.03 (1\ )  & 0.2413 $\pm$ 0.06 (3\ )  \\
  foodtruck       & 0.2384 $\pm$ 0.02 (1)    & 0.1976 $\pm$ 0.01 (4)    & 0.2090 $\pm$ 0.04 (3\ \ ) & 0.1601 $\pm$ 0.01 (7\ \ ) & 0.1692 $\pm$ 0.02 (6\ )  & 0.1280 $\pm$ 0.02 (8)    & 0.1099 $\pm$ 0.01 (9)    & 0.2335 $\pm$ 0.03 (2\ )  & 0.1782 $\pm$ 0.03 (5\ )  & 0.0700 $\pm$ 0.00 (10)   \\
  Water\_quality  & 0.5822 $\pm$ 0.01 (2)    & 0.5836 $\pm$ 0.01 (1)    & 0.4978 $\pm$ 0.01 (6\ \ ) & 0.4527 $\pm$ 0.01 (7\ \ ) & 0.5800 $\pm$ 0.01 (3\ )  & 0.4415 $\pm$ 0.01 (8)    & 0.4239 $\pm$ 0.01 (9)    & 0.5743 $\pm$ 0.02 (4\ )  & 0.5388 $\pm$ 0.01 (5\ )  & 0.0830 $\pm$ 0.02 (10)   \\
  PlantPseAAC     & 0.1760 $\pm$ 0.02 (1)    & 0.1749 $\pm$ 0.02 (2)    & 0.1652 $\pm$ 0.03 (5\ \ ) & 0.1694 $\pm$ 0.03 (4\ \ ) & 0.1708 $\pm$ 0.03 (3\ )  & 0.0968 $\pm$ 0.02 (8)    & 0.0995 $\pm$ 0.01 (6)    & 0.0969 $\pm$ 0.02 (7\ )  & 0.0117 $\pm$ 0.00 (9\ )  & 0.0083 $\pm$ 0.03 (10)   \\
  \hline
  Avg. rank       & \multicolumn{1}{r}{1.38} & \multicolumn{1}{r}{3.15} & \multicolumn{1}{r}{4.27}  & \multicolumn{1}{r}{5.04}  & \multicolumn{1}{r}{5.92} & \multicolumn{1}{r}{6.00} & \multicolumn{1}{r}{6.46} & \multicolumn{1}{r}{6.54} & \multicolumn{1}{r}{7.08} & \multicolumn{1}{r}{9.15} \\
  \hline
\end{tabular}}
\end{table*}

\end{landscape}

The results are presented and analysed in this section. First results from the experiments are presented in Section \ref{subsec:performance_comparison}. Next, in Section \ref{sec:Statsig}, a detailed statistical analysis is performed to understand the overall differences between the algorithms, as well as the per-dataset performance differences.

\subsection{Performance Comparison}\label{subsec:performance_comparison}

Table \ref{tab:big_table} shows the results of the experiments performed. The columns indicate the algorithms and the rows indicate the
datasets. In each cell, the mean and standard deviation label-based macro-averaged
F-Score (higher values are better) across the cross-validation performed are shown. The values in the parenthesis indicate
the relative ranking (lower values are better) of the algorithm with
respect to the corresponding dataset. The last row of Table \ref{tab:big_table}
indicates the overall average ranks of the algorithms compared.

Performance of the compared algorithms in Table \ref{tab:big_table} is interpreted in three ways. Firstly, overall comparison with all the algorithms. Secondly, to compare the individual classifier models compare with ML-KFHE. Finally, and most importantly, to compare ML-KFHE-HOMER with E-HOMER and compare ML-KFHE-CC with ECC directly to understand the effectiveness of the ML-KFHE ensembling.

Table \ref{tab:big_table} shows the overall picture, where ML-KFHE-HOMER attains the best average rank of $1.38$. In fact, ML-KFHE-HOMER attained the top rank for all the datasets, except for \emph{llog} and \emph{Water\_quality} where it got fifth and second rank respectively. E-HOMER attained the second best overall average rank of $3.15$, whereas ML-KFHE-CC attained the third best overall rank of $4.27$. ECC attained the fourth best overall average rank of $5.04$. HOMER-B comes next with overall average rank of $5.92$. The classifiers, CC, RAkEL2, HOMER-K comes next with average ranks of $6.00$, $6.46$ and $6.54$ respectively. RF-PCT was not able to perform well and was ranked $7.08$. The worst performing multi-label classification models in this case was by AdaBoost.MH with an average rank of $9.15$. Similar results using RF-PCT and AdaBoost.MH was also found in \citep{MOYANO201833}.

The difference between E-HOMER and ML-KFHE-HOMER is the aggregation method of the component HOMER classifier models, and ML-KFHE-HOMER has performed better than E-HOMER in all the cases. E-HOMER has similar benefits and drawbacks of a bagged method. E-HOMER models could be trained in parallel as the component classifiers do not depend on each other. Unlike ML-KFHE, E-HOMER gives equal weights to all the component classifiers, good or bad, when combining the models, due to this E-HOMER has a poor predictive performance compared to ML-KFHE-HOMER.

Similarly, in the case ML-KFHE-CC, it performed better than ECC. As in both the cases the difference between E-HOMER, ML-KFHE-HOMER and ECC, ML-KFHE-CC is the combination method of the component classifiers, and all the other processes are kept identical, this demonstrates the effectiveness of ML-KFHE method. Therefore, it can be concluded that ML-KFHE is a better way of ensembling classifiers.

On the other hand, interestingly, E-HOMER has performed better in almost all cases compared to a single HOMER model, as well as having performed better than ECC and ML-KFHE-CC, which demonstrates the effectiveness of ensembling the HOMER method in general. From this experiment it can also be concluded that the KFHE-ML method's overall performance is dependent on the underlying multi-label classifier used, but KFHE-ML method would almost always be able to improve the classification performance when compared to a bagged combination of the underlying classifiers.

\subsection{Statistical Significance Testing and Further Analysis\label{sec:Statsig}}

To further analyse the overall difference of the methods over the different datasets and the differences between per-dataset performances statistical significance tests are performed.

\subsubsection{Multiple Classifier Comparison\label{subsubsec:multiple_classifier_stat}}


\begin{figure}[t!]
\resizebox{\columnwidth}{!}{%
\includegraphics[width=\textwidth]{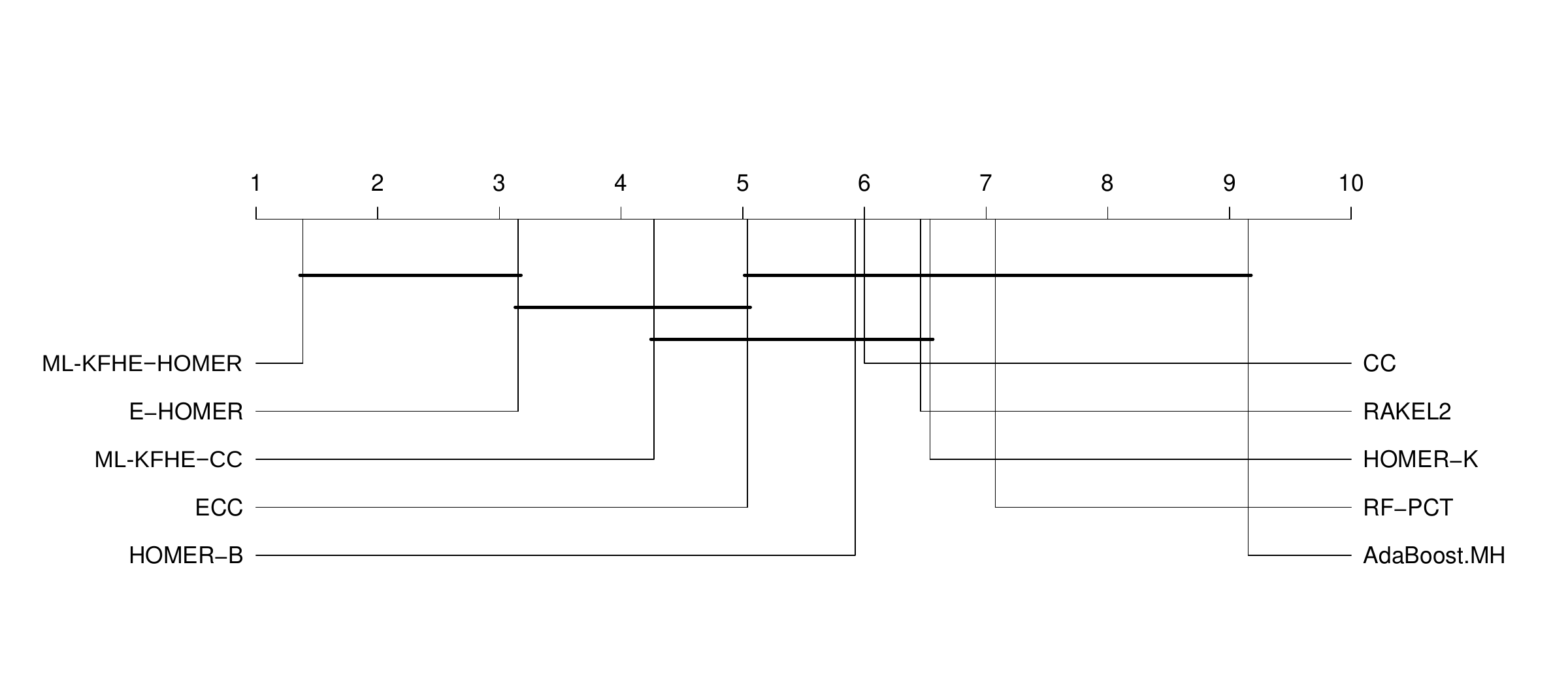}}

\caption{Critical difference plot of Friedman rank sum test with Finner $p$-value correction. The scale indicates the average ranks. The methods which are not connected with the horizontal lines are significantly different with a significance level of $0.05$.\label{fig:friedman}}
\end{figure}

\begin{table}[ht]
\caption{Upper diagonal: win/lose/tie.
Lower diagonal: Results of the Friedman rank test with Finner $p$-value correction. {*} $\alpha=0.1$, {*}{*} $\alpha=0.05$
and {*}{*}{*} $\alpha=0.01$ \label{tab:wlt}}
\centering
\setlength{\tabcolsep}{2pt}
\resizebox{\columnwidth}{!}{%
\begin{tabular}{lllllllllll}
  \hline
                & ML-KFHE-HOMER & E-HOMER    & KFHE-CC    & ECC        & HOMER-B   & CC        & RAkEL2   & HOMER-K  & RF-PCT & AdaBoost.MH \\
  \hline
  ML-KFHE-HOMER &               & 12/1/0     & 13/0/0     & 13/0/0     & 13/0/0    & 12/1/0    & 12/1/0   & 13/0/0   & 12/1/0 & 12/1/0      \\
  E-HOMER       & 0.2166        &            & 10/3/0     & 10/3/0     & 12/1/0    & 10/3/0    & 10/3/0   & 12/1/0   & 12/1/0 & 12/1/0      \\
  ML-KFHE-CC    & 0.0384 **     & 0.4321     &            & 8/5/0      & 9/4/0     & 12/1/0    & 11/2/0   & 9/4/0    & 10/3/0 & 12/1/0      \\
  ECC           & 0.0095 ***    & 0.1985     & 0.6315     &            & 7/6/0     & 10/3/0    & 10/3/0   & 9/4/0    & 9/4/0  & 12/1/0      \\
  HOMER-B       & 0.0007 ***    & 0.0485 **  & 0.2563     & 0.5123     &           & 5/8/0     & 7/6/0    & 9/4/0    & 10/3/0 & 11/2/0      \\
  CC            & 0.0007 ***    & 0.0432 **  & 0.2359     & 0.4812     & 0.9517    &           & 9/4/0    & 7/6/0    & 7/6/0  & 13/0/0      \\
  RAkEL2        & 0.0002 ***    & 0.0184 **  & 0.1319     & 0.3007     & 0.6933    & 0.7139    &          & 7/6/0    & 8/5/0  & 12/1/0      \\
  HOMER-K       & 0.0002 ***    & 0.0163 **  & 0.1195     & 0.2779     & 0.6664    & 0.6933    & 0.9517   &          & 9/4/0  & 11/2/0      \\
  RF-PCT        & 0.0000 ***    & 0.0043 *** & 0.0485 **  & 0.1453     & 0.4321    & 0.4512    & 0.6664   & 0.6933   &        & 11/2/0      \\
  AdaBoost.MH   & 0.0000 ***    & 0.0000 *** & 0.0003 *** & 0.0023 *** & 0.0208 ** & 0.0236 ** & 0.0518 * & 0.0583 * & 0.1453 &             \\
   \hline
\end{tabular}
}
\end{table}

A Friedman rank test was performed with the Finner $p$-value correction \citep{garcia2010advanced}. The results of this evaluation is summarised in Figure \ref{fig:friedman}, where the scale indicates the average ranks and if the methods are not connected with a horizontal line then they are significantly different over different datasets with a significance level of $0.05$. This shows that ML-KFHE-HOMER was significantly better than all the methods except E-HOMER in which case the null hypothesis of Friedman rank test could not be rejected with a significance level of $0.05$. Overall, ML-KFHE-HOMER attained better ranks in all the datasets.

An overall pairwise table of the $p$-values of the Friedman test is shown in Table \ref{tab:wlt}. The lower diagonal of Table \ref{tab:wlt} a value in a cell is the $p$-values of the Friedman rank test with the Finner $p$-value correction for of the corresponding pair of algorithms in the rows and columns. Also, in the upper diagonal of the Table \ref{tab:wlt} each cell has the win/lost/tie count of the algorithm in the corresponding row, over the algorithms in the corresponding column.

\subsubsection{Per-dataset Isolated Pairwise Comparison}

\begin{table}[ht]
\caption{Per-dataset comparison with ML-KFHE-HOMER as the control method of two tailed Wilcoxon's signed rank sum test. The symbols "$\protect\nDownarrow$", "$\protect\ndownarrow$" and "$\protect\ddownarrow$" indicate that the method in the column was \emph{significantly worse} than ML-KFHE-HOMER at a significance level of $0.01$, $0.05$ and $0.1$, respectively on the specific dataset. Similarly, "$\protect\nUparrow$", "$\protect\nuparrow$" and "$\protect\duparrow$" symbols indicate that the method in the column was \emph{significantly better} than ML-KFHE-HOMER with a significance level of $0.01$, $0.05$ and $0.1$, respectively on the specific dataset. The "$.$" indicates that the null hypothesis for the mentioned two tailed Wilcoxon's signed rank sum test could not be rejected with any significance level.\label{tab:kfhe_homer_vs_all}}
\centering
\resizebox{1.0\columnwidth}{!}{%
\begin{tabular}{rccccccccc}
  \hline
                 &                                                    \multicolumn{9}{c|}{ML-KFHE-HOMER vs.}                                                        \\
  \hline
                 & ML-KFHE-CC    & E-HOMER           & ECC           & HOMER-B       & CC            & RAkEL2        & HOMER-K       & RF-PCT        & AdaBoost.MH   \\
  \hline
  flags          & $\nDownarrow$ & .                 & $\nDownarrow$ & .             & $\nDownarrow$ & $\nDownarrow$ & $\nDownarrow$ & $\ndownarrow$ & $\nDownarrow$ \\
  yeast          & $\nDownarrow$ & $\nDownarrow$     & $\nDownarrow$ & $\nDownarrow$ & $\nDownarrow$ & $\nDownarrow$ & $\nDownarrow$ & $\nDownarrow$ & $\nDownarrow$ \\
  scene          & $\ddownarrow$ & $\ddownarrow$     & $\nDownarrow$ & $\nDownarrow$ & $\ndownarrow$ & .             & $\nDownarrow$ & $\nDownarrow$ & $\nDownarrow$ \\
  emotions       & $\nDownarrow$ & .                 & $\ndownarrow$ & .             & $\ndownarrow$ & .             & $\ddownarrow$ & $\nDownarrow$ & $\nDownarrow$ \\
  medical        & $\ndownarrow$ & $\ddownarrow$     & .             & $\nDownarrow$ & $\ndownarrow$ & $\nDownarrow$ & $\nDownarrow$ & $\nDownarrow$ & $\nDownarrow$ \\
  enron          & $\ddownarrow$ & .                 & $\ddownarrow$ & $\nDownarrow$ & $\nDownarrow$ & $\nDownarrow$ & $\nDownarrow$ & $\nDownarrow$ & $\nDownarrow$ \\
  birds          & $\ndownarrow$ & .                 & $\ndownarrow$ & $\ndownarrow$ & $\ndownarrow$ & $\nDownarrow$ & $\nDownarrow$ & $\nDownarrow$ & $\nDownarrow$ \\
  genbase        & .             & $\nDownarrow$     & .             & $\nDownarrow$ & .             & $\ddownarrow$ & $\nDownarrow$ & $\nDownarrow$ & $\nDownarrow$ \\
  cal500         & $\ddownarrow$ & .                 & $\ndownarrow$ & $\nDownarrow$ & $\nDownarrow$ & $\nDownarrow$ & $\nDownarrow$ & $\ndownarrow$ & $\nDownarrow$ \\
  llog           & .             & .                 & .             & .             & $\duparrow$   & .             & $\nDownarrow$ & .             & .             \\
  foodtruck      & $\ndownarrow$ & $\nDownarrow$     & $\nDownarrow$ & $\nDownarrow$ & $\nDownarrow$ & $\nDownarrow$ & .             & $\nDownarrow$ & $\nDownarrow$ \\
  Water\_quality & $\nDownarrow$ & .                 & $\nDownarrow$ & .             & $\nDownarrow$ & $\nDownarrow$ & .             & $\nDownarrow$ & $\nDownarrow$ \\
  PlantPseAAC    & $\ddownarrow$ & .                 & .             & .             & $\nDownarrow$ & $\nDownarrow$ & $\nDownarrow$ & $\nDownarrow$ & $\nDownarrow$ \\
   \hline
\end{tabular}}
\end{table}

\begin{table}[h!]
\caption{Per-dataset comparison with ML-KFHE-CC as the control method of two tailed Wilcoxon's signed rank sum test. The symbols "$\protect\nDownarrow$", "$\protect\ndownarrow$" and "$\protect\ddownarrow$" indicate that the method in the column was \emph{significantly worse} than ML-KFHE-CC at a significance level of $0.01$, $0.05$ and $0.1$, respectively on the specific dataset. The "$\protect\nUparrow$", "$\protect\nuparrow$" and "$\protect\duparrow$" symbols indicate that the method in the column was \emph{significantly better} than ML-KFHE-CC with a significance level of $0.01$, $0.05$ and $0.1$, respectively on the specific dataset. The "$.$" indicates that the null hypothesis for the mentioned two tailed Wilcoxon's signed rank sum test could not be rejected with any significance level.\label{tab:kfhe_cc_vs_all}}
\centering
\resizebox{1.0\columnwidth}{!}{%
\begin{tabular}{rccccccccc}
  \hline
                 &                                                    \multicolumn{9}{c|}{ML-KFHE-CC vs.}                                                        \\
  \hline
                 & ML-KFHE-HOMER & E-HOMER       & ECC           & HOMER-B       & CC            & RAkEL2        & HOMER-K       & RF-PCT        & AdaBoost.MH   \\
  \hline
  flags          & $\nUparrow$   & $\nuparrow$   & .             & $\nuparrow$   & .             & $\ndownarrow$ & $\duparrow$   & .             & $\ddownarrow$ \\
  yeast          & $\nUparrow$   & $\ndownarrow$ & $\ndownarrow$ & $\nDownarrow$ & $\ndownarrow$ & $\nDownarrow$ & $\nDownarrow$ & $\nDownarrow$ & $\nDownarrow$ \\
  scene          & $\duparrow$   & .             & $\ndownarrow$ & $\nDownarrow$ & $\ndownarrow$ & .             & $\nDownarrow$ & $\nDownarrow$ & $\nDownarrow$ \\
  emotions       & $\nUparrow$   & .             & .             & $\duparrow$   & .             & .             & .             & .             & $\nDownarrow$ \\
  medical        & $\nuparrow$   & .             & .             & $\nDownarrow$ & .             & .             & $\nDownarrow$ & $\nDownarrow$ & $\nDownarrow$ \\
  enron          & $\duparrow$   & $\duparrow$   & .             & $\nDownarrow$ & $\nDownarrow$ & $\nDownarrow$ & $\nDownarrow$ & $\nDownarrow$ & $\nDownarrow$ \\
  birds          & $\nuparrow$   & .             & .             & $\ddownarrow$ & $\ddownarrow$ & $\ndownarrow$ & $\ndownarrow$ & $\nDownarrow$ & $\nDownarrow$ \\
  genbase        & .             & $\ddownarrow$ & .             & $\nDownarrow$ & .             & .             & $\ndownarrow$ & $\nDownarrow$ & $\nDownarrow$ \\
  cal500         & $\duparrow$   & $\nuparrow$   & .             & $\nDownarrow$ & $\nDownarrow$ & $\nDownarrow$ & $\nDownarrow$ & $\ddownarrow$ & $\nDownarrow$ \\
  llog           & .             & .             & .             & .             & $\nuparrow$   & $\nuparrow$   & $\nDownarrow$ & $\duparrow$   & .             \\
  foodtruck      & $\nuparrow$   & .             & $\nDownarrow$ & $\ndownarrow$ & $\nDownarrow$ & $\nDownarrow$ & $\nuparrow$   & $\ndownarrow$ & $\nDownarrow$ \\
  Water\_quality & $\nUparrow$   & $\nUparrow$   & $\nDownarrow$ & $\nUparrow$   & $\nDownarrow$ & $\nDownarrow$ & $\nUparrow$   & $\nUparrow$   & $\nDownarrow$ \\
  PlantPseAAC    & $\duparrow$   & .             & .             & .             & $\nDownarrow$ & $\nDownarrow$ & $\nDownarrow$ & $\nDownarrow$ & $\nDownarrow$ \\
   \hline
\end{tabular}}
\end{table}

In the previous section the overall performance of the algorithms over different datasets were analysed. Now, how different the performance (label-based macro-averaged F-Score) on each individual datasets are, will be analysed. To understand if ML-KFHE-HOMER and ML-KFHE-CC did attain significantly different (better or worse) results than the other methods for each dataset, a two-tailed paired Wilcoxon's signed rank sum test \citep{garcia2010advanced} was performed over the folds of each cross-validation experiment. Two tests were done. First, ML-KFHE-HOMER was set as the control method and compared to the other methods per dataset. Next, ML-KFHE-CC was set as the control method and compared with the other methods per dataset.


The result from the first experiment where KFHE-ML-HOMER is the control method, is shown in Table \ref{tab:kfhe_homer_vs_all} and the results of the second experiment with ML-KFHE-CC is the control method, is shown in Table \ref{tab:kfhe_cc_vs_all}. This means, when for example the \emph{yeast} dataset, there are a set of scores from each fold of the 2 times 5 fold crossvalidation experiment (total 10 label-based macro-averaged F-Scores) for ML-KFHE-HOMER and ECC each. The arrows indicate the result of comparing these two sets of scores using the Wilcoxon's signed rank sum test indicating that ECC was significantly worse than ML-KFHE-HOMER.

The symbols "$\nDownarrow$", "$\ndownarrow$" and "$\ddownarrow$" in Table \ref{tab:kfhe_homer_vs_all} and \ref{tab:kfhe_cc_vs_all} indicate that the method in the column was \emph{significantly worse} than the control method in the corresponding table at a significance level of $0.01$, $0.05$ and $0.1$, respectively on the specific dataset. The $\nUparrow$, $\nuparrow$, "$\duparrow$" symbols indicate that the method in the column was \emph{significantly better} than control method in the corresponding table with a significance level of $0.01$, $0.05$ and $0.1$, respectively on the specific dataset. The "$.$" symbol in both Tables \ref{tab:kfhe_homer_vs_all} and \ref{tab:kfhe_cc_vs_all} indicates that the null hypothesis for the mentioned two tailed Wilcoxon's signed rank sum test could not be rejected with any significance level. For example, in Table \ref{tab:kfhe_homer_vs_all} (with ML-KFHE-HOMER as the control method), ECC was significantly worse than ML-KFHE-HOMER with a significance level of $0.01$ in the case of the following datasets: \emph{flags}, \emph{yeast}, \emph{scene}, \emph{foodtruck}, \emph{Water\_quality}. In the case of \emph{emotions}, \emph{birds} and \emph{cal500} ECC was significantly worse with a level of $0.05$ and in the case of \emph{enron} it was significantly worse with a level of $0.1$. Similarly, in Table \ref{tab:kfhe_cc_vs_all} (with ML-KFHE-CC as the control method), ECC performed significantly worse with a level of $0.01$ on \emph{foodtruck} and \emph{Water\_quality} datasets, and significantly worse with a level of $0.05$ in the case of \emph{yeast} and \emph{scene} dataset.

Some interesting patterns can be observed from Table \ref{tab:kfhe_homer_vs_all} and \ref{tab:kfhe_cc_vs_all}. It is clear that the variant ML-KFHE-HOMER was significantly better in the case of almost all the datasets compared to ML-KFHE-CC, which clearly indicates that ML-KFHE-HOMER is the better variant. Considering ML-KFHE-HOMER vs E-HOMER and HOMER-K and HOMER-B in Table \ref{tab:kfhe_homer_vs_all} it can be seen that ML-KFHE-HOMER was able to significantly improve on many datasets and never got a worse rank (except in the case of \emph{llog}) as shown in Table \ref{tab:big_table}. Therefore, ML-KFHE-HOMER is always better than the component classifiers as well as a better choice than bagging aggregation approach (E-HOMER). For the other methods, ML-KFHE-HOMER was able to significantly improve the label-based macro-averaged F-Scores in almost all the datasets. Interestingly, CC was significantly better than ML-KFHE-HOMER in the case of \emph{llog} with a significance level of $0.1$.

Considering ML-KFHE-CC as the control method in Table \ref{tab:kfhe_cc_vs_all} it can be seen that ECC has performed significantly worse in the case of \emph{foodtruck} and \emph{Water\_quality} with a significance level of $0.01$ and significantly worse in the case of \emph{yeast} and \emph{scene} with a significance level of $0.05$. Although ECC performed better in some cases than ML-KFHE-CC (table \ref{tab:big_table}), but for those datasets the null hypotheses could not be rejected in any of the significance levels in this experiment. Also, it is clear that ML-KFHE-CC has performed significantly better in the case of almost all the datasets compared to CC (except \emph{llog}). This shows that ML-KFHE-CC is most of the cases a better ensemble technique compared to ECC, but not as good as ML-KFHE-HOMER.

It must be emphasised that the Wilcoxon's signed rank sum test \emph{cannot}
be used to perform multiple classifier comparison without introducing
Type I error (rejecting the null hypothesis when it cannot be rejected),
as it does not control the Family Wise Error Rate (FWER) \citep{garcia2010advanced} in the above analysis.
Therefore, each pair from this experiment should
\emph{only} be interpreted in isolation from any other algorithms. Multiple classifier comparison is done in Section \ref{subsubsec:multiple_classifier_stat}.

Overall, it can be concluded that the ML-KFHE-HOMER improves the label-based macro-averaged F-Scores significantly in almost all the datasets when compares to any of the algorithms. E-HOMER (also introduced in this work) is an effective technique as well, but ML-KFHE-HOMER almost always performs better than E-HOMER, thus demonstrating the effectiveness of the KFHE framework for ensembling compared to a bagged method for ensemblig. ML-KFHE-CC also performed well, but not as good as ML-KFHE-CC. When compared to ECC, ML-KFHE-CC was able to improve performance of the model on several datasets, but the difference between ECC and ML-KFHE-CC was not as large as what it is between E-HOMER and ML-KFHE-HOMER. Also, the training time growth of HOMER is much faster than CC, which makes HOMER scailable. These leads to the conclution that ML-KFHE-HOMER is a much superior variant.

\section{Conclusions and Future Work\label{sec:Conclusion-and-Future}}

In multi-label literature there are several ensemble methods which are mostly based on bagging or majority voting methods \citep{MOYANO201833}, which perform well. As in multi-class classification, boosting methods generally performs much better than a single classifier model. But boosting or boosting-like methods are rarely explored in the multi-label literature.

This work introduces a multi-label classification method, ML-KFHE, that exploits the sensor fusion properties of the Kalman filter, used in the Kalman Filter-based Heuristic Ensemble (KFHE). Given the nature of the algorithm, effectively, this falls in the middle of boosting and bagging. ML-KFHE views the ensemble classifier model to be trained as a state to be estimated and does so using a Kalman filter that combines multiple noisy measurements, where each measurement is a trained classifier and the noise is its related classification error.

In ML-KFHE, the sensor fusion properties of the Kalman filter is used to aggregate multiple and diversely trained HOMER or CC models. The method ensembles multiple HOMER or CC models trained on weighted samples of a training dataset and using different hyperparameter settings. The KFHE framework combines these models based on the classification error of the HOMER or CC models. Summary of the findings are as follows


\begin{itemize}
  \item ML-KFHE was able to perform consistently better than its component classifiers.
  \item The aggregation method of ML-KFHE using the Kalman filter is more effective than existing and common methods of bagging-like combination as in ECC or E-HOMER, therefore showing the effectiveness of the KFHE framework.
  \item The ML-KFHE-HOMER variant performed better than ECC, RAkEL and the other multi-label ensemble methods evaluated.
\end{itemize}

ML-KFHE might converge too fast if the several component classifiers in a sequence have high bias but low variance. This can result in ML-KFHE to perform suboptimally. Presently the random hyperparameter selection of the component classifiers introduce the diversity to stop this happening. Also, in the later iterations of ML-KFHE when the uncertainty of the ensemble is reducing (Kalman gain reduces), if a new component classifier model is found (due to a randomly selected good hyperparameter) to be more accurate, due to the lower uncertainty of the ensemble, the new more accurate measurement may not be incorporated into the model due to a lower Kalman gain value. To stop the method converging too fast, process noise or a slowdown mechanism can be introduced, which may improve performance in some cases where the Kalman gain becomes $1$ (one of the measurements was perfect). Also, it would also be interesting to formally compare the training and prediction runtime of the different methods.

Also, In the future a \emph{per-label} version of ML-KFHE could be explored, where instead of the combination of multiple labels using one Kalman gain, per-label Kalman gains will be maintained. The present algorithm does not have a time update step, which can also be introduced and studied.


\section*{Data Availability Statement}

The datasets generated during and/or analysed during the current study are available in the following repositories: http://mulan.sourceforge.net/datasets-mlc.html, http://www.uco.es/kdis/mllresources/ .

\section*{Declarations}
\textbf{Conflict of interests:} The authors have no financial or non-financial conflicts of interests to disclose.

\section*{Acknowledgements}

This publication has emanated from research conducted with the financial support of Science Foundation Ireland under Grant number [16/RC/3835]. For the purpose of Open Access, the author has applied a CC BY public copyright licence to any Author Accepted Manuscript version arising from this submission. The authors also would like to acknowledge Dr. Derek Greene for valuable inputs to improve the quality of the draft.

\bibliography{kfhe_ml}


\end{document}